\documentclass[letterpaper]{article} % DO NOT CHANGE THIS
\usepackage{aaai23}
\usepackage{times}  % DO NOT CHANGE THIS
\usepackage{helvet}  % DO NOT CHANGE THIS
\usepackage{color}
\usepackage{makecell}
\usepackage{courier}  % DO NOT CHANGE THIS
\usepackage[hyphens]{url}  % DO NOT CHANGE THIS
\usepackage{graphicx}
\def\UrlFont{\rm}  % DO NOT CHANGE THIS
\usepackage{natbib}  % DO NOT CHANGE THIS AND DO NOT ADD ANY OPTIONS TO IT
\usepackage{caption} % DO NOT CHANGE THIS AND DO NOT ADD ANY OPTIONS TO IT
\usepackage{algorithm}
\usepackage{algorithmic}
\usepackage{amsmath}
\usepackage{multicol}
\usepackage{multirow}

\usepackage{newfloat}
\usepackage{listings}
\DeclareCaptionStyle{ruled}{labelfont=normalfont,labelsep=colon,strut=off} % DO NOT CHANGE THIS
\floatstyle{ruled}
\newfloat{listing}{tb}{lst}{}
\floatname{listing}{Listing}
\title{Complex Dynamic Neurons Improved Spiking Transformer Network \\for Efficient Automatic Speech Recognition}
\author{
    Minglun Han\textsuperscript{\rm 1,2}\equalcontrib,
    Qingyu Wang\textsuperscript{\rm 1,2}\equalcontrib,
    Tielin Zhang\textsuperscript{\rm 1,2}\equalcontrib
    \footnote{Corresponding authors.},
    Yi Wang\textsuperscript{\rm 4},
    Duzhen Zhang\textsuperscript{\rm 1,2},
    Bo Xu\textsuperscript{\rm 1,2,3}\footnotemark[2]
}
\affiliations{
    \textsuperscript{\rm 1} Institute of Automation, Chinese Academy of Sciences, Beijing, China\\
    \textsuperscript{\rm 2} School of Artificial Intelligence, University of Chinese Academy of Sciences, Beijing, China\\
    \textsuperscript{\rm 3} Center for Excellence in Brain Science and Intelligence Technology, Chinese Academy of Sciences, Shanghai, China\\
    \textsuperscript{\rm 4} School of Artificial Intelligence, Jilin University, Changchun, China\\
    $\{$tielin.zhang, xubo$\}$@ia.ac.cn
}

\begin{document}

\maketitle

\begin{abstract}
The spiking neural network (SNN) using leaky-integrated-and-fire (LIF) neurons has been commonly used in automatic speech recognition (ASR) tasks. However, the LIF neuron is still relatively simple compared to that in the biological brain. Further research on more types of neurons with different scales of neuronal dynamics is necessary. Here we introduce four types of neuronal dynamics to post-process the sequential patterns generated from the spiking transformer to get the complex dynamic neuron improved spiking transformer neural network (DyTr-SNN). We found that the DyTr-SNN could handle the non-toy automatic speech recognition task well, representing a lower phoneme error rate, lower computational cost, and higher robustness. These results indicate that the further cooperation of SNNs and neural dynamics at the neuron and network scales might have much in store for the future, especially on the ASR tasks.
\end{abstract}

\section{Introduction}

In recent years, the spiking neural network (SNN) has received extensive attention for its remarked lower computational cost on various machine learning tasks, including but not limited to spatial image classification, temporal auditory recognition, event-based video recognition, and reinforcement-learning based continuous control \cite{tang2021deep}. This progress in SNNs is contributed partly by the mathematical optimization algorithms borrowed from the artificial neural network (ANN), e.g., the approximate gradient in backpropagation (BP), various types of loss definition and regression configuration, and more importantly, by some key computational modules inspired from the biological brain, e.g., the receptive-field-like convolutional circuits, self-organized plasticity propagation \cite{RN767}, heterogeneity of artificial neurons for the robust computation \cite{RN778}, and other multi-scale inspiration from the single neuron or synapse to the network or cognitive functions. 

However, many key modules in SNNs are relatively simple, which stops their further improvement of accuracy, robustness, and efficiency. For example, the leaky-integrated-and-fire (LIF) neuron is commonly used in conventional SNNs, mainly caused of their simplicity and efficiency during network learning. Similar to it, integrated and fire (IF) neuron is also generally used in many ANN-to-SNN conversion algorithms, partly caused by the mathematical equivalence of activation functions between IF in SNN and rectified linear unit (ReLU) in ANN.

Neuronal dynamics and neuronal heterogeneity are important features for efficient sequential information processing \cite{RN448} and robust computation \cite{RN778}. The spiking neuron with different dynamics in the biological brain is usually considered the basic building block to support cognitive functions at higher scales. More biologically plausible features, such as network topology and plasticity-based learning algorithms, are also very important but will not be further discussed in this study. The SNN is considered the third-generation ANN \cite{RN670}, guiding the conventional ANN to the goal of biologically plausible machine intelligence. The SNN could incorporate spatial and temporal information at multi scales, making them efficient in many cognitive functions, including but not limited to the spatial or temporal information sensation, working memory formation with various dynamics, and complex decision-making based on short or long-term memory. Most of these key features are inspired by the biological counterpart brain and might be the key to balancing the accuracy and computational cost. 

Even though many efforts have been given to integrate SNNs, ANNs, and biological discovery in different manners from both software and hardware perspectives, it is still a big challenge to combine them naturally by resolving some conflict features of them, including but not limited to different types of learning neurons, architectures, and learning algorithms. Some researchers try to combine ANN and SNN in a unified framework by freely setting spike-based and rate-based neurons towards the artificial general intelligence under the Tianjic neuromorphic computation platform \cite{RN436}. The Spaun architecture integrates global rate-based routing algorithms and local spiking neurons to build functional brain regions to support more than eight cognitive functions \cite{RN548}. Some SNNs try to get around the non-differential optimization challenge by training a deep ANN first and then converting it into a spike version for achieving high accuracy \cite{RN849}. Then both gradient-based and plasticity-based algorithms have been proposed to train SNN for different machine-learning tasks. The SNN using gradient-based algorithms could usually achieve relatively higher performance, which we primarily used in this paper. It is generally accepted that, on the one hand, directly incorporating spiking neurons into ANNs might make the gradient in BP non-precise, and on the other hand, the complex inner dynamics of spiking neurons would make the convergence speed slower. However, the integration of ANNs and SNNs is necessary because gradient learning in ANN has been verified as efficient, and many key features in SNNs borrowed from natural neural networks might be the key to reaching biology-like computational efficiency, flexibility, and robustness.

Hence, this paper integrates some key features from ANNs and biology to improve SNN. The main contributions of this paper can be concluded in the following parts.

\begin{itemize}
  \item First, four spiking neurons with complex dynamics are designed and used for efficient temporal information processing in SNNs, containing the vanilla 1st-order dynamics, 1st-order dynamics with adaptive threshold, 2nd-order dynamics, and double-neuron dynamics with inhibitory neurons. 
  \item Second, we borrow the transformer module in ANN and revise it as a spiking transformer to reduce the computational cost, without seriously affecting the classification accuracy. Furthermore,  an autoregression network is also used for the signal decoder.
  \item Third, a benchmark automatic speech recognition (ASR) task is used to test the proposed algorithm on its accuracy, efficiency, and robustness, where the moderate LJSpeech dataset contains sequential spoken phonemes and words in a full sentence.
  \item Fourth, after analyzing, we find that the inter-spike-interval in SNNs can naturally represent the phoneme chunk boundaries, representing handling the implicit rhythm of long and short phonemes by heterogeneous neuronal dynamics.
\end{itemize}

\section{Related work}

% For ANNs, many algorithms try to find inspiration from interdisciplinary integration. Borrowing some new ideas from biological intelligence is a quick, easy and short path. Inspired by lateral inhibition, some lateral connections were designed in conventional convolutional neural networks and have been found effective in the robust computation \cite{RN494}. A self-backpropagation learning algorithm was proposed to guide the synaptic modification for reaching higher accuracy and lower computational cost on both ANNs and SNNs \cite{RN767}.

For SNNs, two main research directions are formalized caused of different goals. One is to understand the biological brain better through computer simulation. A spiking-version adaptive dynamic processing algorithm is designed for the discrete-time optimization with the Poisson process \cite{RN804}. The two-scale clocks in convolutional spiking neurons are unified and well-tuned with biologically plausible reward propagation \cite{RN765}. Another is to achieve the highest performance to move artificial intelligence further. Population coding at the network scale and spike coding at the neuronal scale were integrated and exhibited higher performance than ANNs on continuous-control tasks during reinforcement learning \cite{RN798}. An e-prop algorithm was proposed that incorporated the spiking units and target propagation for reaching higher performance on both speech recognition and reinforcement learning \cite{RN557}.

For spatial or temporal information processing, many methods \cite{mueller2021spiking} have been proposed in ANNs. While in SNNs, different types of meta-dynamic neurons were designed to support the general spatial and temporal information processing \cite{RN478}. It has been reported that multisensory integration exists in a single neuron \cite{RN302}, synapse \cite{RN309}, or different scales of the brain \cite{RN304}. The 3-node network motif was used to represent the feature of biological topology and verified efficiency in multisensory integration after copying these biological features to the SNNs \cite{RN801}. The dynamic firing threshold was simulated with one-order dynamics and exhibited power on temporal information processing \cite{RN766}. The new-general dynamic vision sensor was designed to process spike-based event signals, which showed the efficient ability to process temporal information with an extremely high frame rate than the conventional frame-based cameras \cite{RN308}. The efficient spatially temporal information processing of SNNs has been verified on many biological and realistic artificial intelligence tasks \cite{RN307}. 

\section{Methods}

\begin{figure*}[htbp]
\centering
\includegraphics[width=17cm]{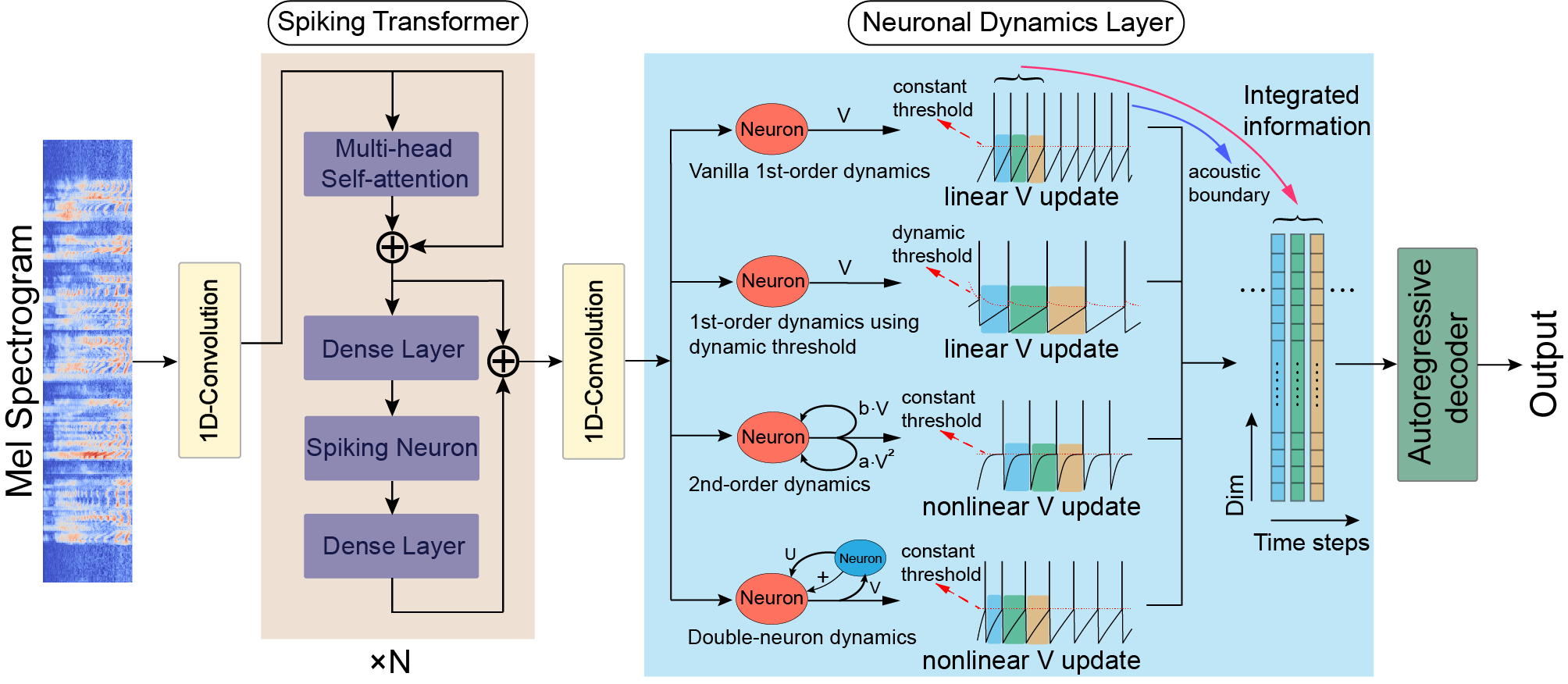}
\caption{The overall architecture of the proposed DyTr-SNN.}
\label{fig_overall}
\end{figure*}

The overview of our DyTr-SNN is presented in Figure \ref{fig_overall}. First, two layers with 1D-convolution kernels are designed to conduct down-sampling and acoustic modeling of raw speech features. Second, N layers with spiking transformers are designed for spatially temporal information encoding. Each transformer contains 6 blocks, including muti-head self-attention, first residual-connection, dense layer (fully connected layer, FC), spiking neuron, dense layer (FC), and second residual-connection. Third, one layer with 1D-convolutional kernels is designed to learn context information. Fourth, an FC layer is designed to calculate a sequence of current signals. Fifth, the one layer with complex neuronal dynamics is designed to detect phone chunk
boundaries and integrate information, where four main types of neurons are introduced. Finally, the output layer with an auto-regressive decoder calculates the final output.
%the network input using sampled auditory stimuli at 20 kHz and processed with raw speech features; 

The heterogeneity of neuronal dynamics is the most important module in this study, where the spikes could be detected and represent the phone chunk boundary. 

\subsection{The encoder layer with spiking transformer}

The transformer-encoder layer receives the raw speech features as network input (see Experiments for more details) and generates high-dimensional acoustic information representations. The whole procedure contains the following steps. 

First, we use a convolutional layer as the front end, which utilizes a convolution layer with stride 2 to conduct temporal down-sampling of raw speech features. This process can reduce the length of temporal feature sequences by using learned kernel filters to reduce the memory cost for efficient training.

Second, we use a spiking transformer to improve parallel computing effectively. The raw transformer contains artificial neurons. It uses scaled dot-product attention to find the correlation between sub-elements of the input sequence. It has three inputs: query ($Q$), key ($K$), and value ($V$). Since the speech sequence in the ASR task contains complex acoustic information at the phoneme level as well as semantic information at the word level, in order to map the sequence to multiple high-level spaces and obtain rich information on different scales as much as possible, the selection of multi-head attention layer is necessary. Then the calculated multiple self-attention are combined to reach a final output using a full connection layer.

Recently, many spiking transformer architectures have been proposed. They highlight spatio-temporal feature fusion \cite{zhang2022spiking1}, or make ANN-to-SNN conversion \cite{mueller2021spiking}. Here we first replace the ReLU activation function with integration-and-fire (IF) neurons (i.e., spiking neurons with simple 1st-order neuronal dynamics) due to their similarities and then train the SNN transformer directly. The IF is the simplest type of spiking neuron model and could be defined as the following Equation:

\begin{equation}
\left\{
\begin{array}{lr}
\vspace{1ex}
C_m \frac{dV_{k,t}}{dt}= I_k &  \\
V_{k,t}=V_{reset}, Spike=1 \quad if(V_{k,t}>V_{th}) &
\end{array}
\right.
\text{,}
\end{equation}

where $V_{k,t}$ is the membrane potential for the neuron $k$ at time $t$, $C_m$ is the membrane capacitance, $V_{th}$ is the firing threshold, $V_{reset}$ is the reset potential, $Spike$ is the firing flag. As the current $I_{k}$ inputs into the neuron $k$ with time steps, its membrane potential gradually increases and finally reaches the threshold, meanwhile the neuron will fire a spike and then be reset to $V_{reset}$, waiting for a next firing process. The relationship between fire rate in ANN ($R_i^{ANN}$) and spike trains (generated with IF neuron, $V_i^{SNN}$) in SNN is shown as the following Equation:

\begin{equation}
R(t)=R_i^{ANN} R_{max}- \frac{V_i^{SNN}}{t V_{th}}
\text{,}
\end{equation}

where the fire rate after conversion is represented as the temporal firerate $R(t)$, $R_i^{ANN}$ is the fire rate value of the neuron $i$ after the ReLU activation, $R_{max}$ is the maximal firing rate of the ANN, $V_i^{SNN}$ is the membrane potential, $t$ is the time step.

\subsection{The simple IF neuron for the boundary detection}

The spiking neuron encoder is the most important part after the spiking transformer. The IF neuron is used to process temporal acoustic, with the function of membrane potential accumulation and spike firing in the neuron model, to locate acoustic boundaries and dynamic information integration. This special process includes the following 3 steps:

First, a sequence of current signals  $I=(I_1, I_2, I_3...)$ is used to describe the amount of acoustic information calculated from the sequentially input $h=(h_1, h_2, h_3...)$ during the spiking transformer layer. A 1-dimensional convolution is used to capture the local dependency, and then a Sigmoid function is used to non-linear normalize the signal to scalar values between 0 and 1. 

Second, the current signal $I$ will stimulate a neuron to generate spikes after a period of membrane-potential accumulation. For example, at the time step $t$ during learning, a function $F(I_{k,t},V_{k,t-1},...)$ of current value $I_{k,t}$ in the current sequence, as well as some other variables such as membrane potential $V_{k,t-1}$ at previous step $t-1$, will be calculated first and then added to the previous membrane potential $V_{k,t-1}$ to get the new accumulated membrane potential $V_{k,t}$. This procedure could be described in the following Equation:

\begin{equation}
V_{k,t}=V_{k,t-1}+F(I_{k,t},V_{k,t-1},...) 
\text{,}
\end{equation}

where if $V_{k,t}$ is less than a given threshold $V_{th}$, it will still be used as the previous potential for the next step $t+1$, or otherwise, it will be set as $V_{k,t} -1$, representing a leakage of membrane potential due to spike firing. The value of 1 here is unimportant since other parameters (e.g, the synaptic weights) could adapt to it during learning. It can also be described as the following Equation, where the time constant $\tau$ is set to 1 for simplicity.

\begin{equation}
\tau \frac{dV_{t}}{dt}=F(I_{k,t},V_{k,t-1},...) 
\text{.}
\end{equation}

Notably, various possible expressions here are included in function $F$, each representing a kind of neuronal dynamics. This study will introduce four main dynamic neurons in the next subsection.

Hence, we can now determine whether an acoustic boundary is located by observing whether there is a spike generated from the dynamic neurons. If no boundary was located, we could update the current integrated state with the function of $S_{t}=S_{t-1}+V_{t} \cdot h_t$ for the next step $t+1$. If $V_{t}$ was greater than $V_{th}$, it means an acoustic boundary is located, and $S_t$ should be sent to the next-step decoder module as the integrated acoustic information corresponding to the current token. All these calculations would be repeated until the network is convergent and the final acoustic representations sequence is generated at network output.

\subsection{Spiking neurons with different neuronal dynamics}

In this study, four types of neurons are introduced to discuss their ability for acoustic boundary identification further: (1) the vanilla 1st-order dynamics, which uses a predefined firing threshold and is similar to that in the IF neuron; (2) the 1st-order dynamics with an adaptive threshold; (3) the 2nd-order dynamics; (4) the double-neuron dynamics using an additional neuron for the feedback inhibition.

\subsubsection{The vanilla 1st-order dynamics}

Neurons with vanilla 1st-order dynamics are similar to the IF neurons, which have already been proven effective in many types of research \cite{linhao2018cif,HanDZX21,HanDLCZMX22}. The description of the vanilla neuron is shown in the following Equation (\ref{equa_IF}):

\begin{equation}
\frac{dV_{k,t}}{dt}=I_{k,t}
\text{,}
\label{equa_IF}
\end{equation}

where $I_{k,t}$ is the input current to the neuron $k$, and $V_{k,t}$ is the membrane potential. It describes that the membrane update is only related to the value of the input current, indicating a nearly linear process of membrane potential accumulation. A schematic diagram depicting the membrane potential under the vanilla 1st-order dynamics is also shown in Figure \ref{fig_overall}.

\subsubsection{The 1st-order dynamics using dynamic threshold}

Unlike the IF neuron, where the firing threshold is a predefined static threshold value, the neurons with 1st-order dynamics using adaptive threshold run further to improve their dynamic complexity by introducing this special dynamic threshold which is changed at each step. The additional computation of the dynamic threshold is shown as the following Equation (\ref{equa_IF_dy_thre}):

\begin{equation}
V_{th,t}= \alpha V_{th,t-1}  + Spike \cdot \Delta h
\label{equa_IF_dy_thre}
\end{equation}

where $V_{th,t}$ is the firing threshold at time step $t$, and $V_{th,t-1}$ is the threshold at the next step $t-1$,  $\alpha$ is the decay coefficiency, $\Delta h$ is the jump coefficiency once receiving the neuronal firing flag $Spike$. This function describes that the neuronal threshold will get a jump to increase the difficulty of firing another spike in a short time. This function could also be considered temporal masking in the time domain, to promote firing for sparse spikes or, on the contrary, decrease firing for dense spikes. This mechanism could provide a richer dynamic process for inner neurons. A schematic diagram depicting the membrane potential under the 1st-order dynamics using adaptive threshold is also shown in Figure \ref{fig_overall}.

\subsubsection{The 2nd-order dynamics}

We further discuss some neurons with higher-order dynamics, which have been proven effective in reinforcement learning \cite{RN798}. It means the membrane potential might have more than one equilibrium state. In this paper, we introduce neurons with 2nd-order dynamics, shown as the following Equation (\ref{equa_2nd_order}):

\begin{equation}
\frac{dV_{k,t}}{dt}=a\cdot V_{k,t}^2 + b\cdot V_{k,t} + c \cdot I_{k,t}
\text{,}
\label{equa_2nd_order}
\end{equation}

$V_{k,t}^2$ and $V_{k,t}$ are membrane potentials with different degrees of dynamics for the neuron $k$ at time $t$, $a$, $b$ and $c$ are the hyperparameters to represent the corresponding coefficients of $V_{k,t}$. As a result, the dynamic membrane potential will be attracted or nonstable at some points given a very small input and will have a non-linear growth given a bigger input. A schematic diagram depicting the membrane potential under the 2nd-order dynamics is also shown in Figure \ref{fig_overall}.

\subsubsection{The double-neuron dynamics}

Double-neuron dynamics means an excitatory neuron could receive a self-inhibition as input from another inhibitory neuron. The dynamic changes of membrane potential for both excitatory ($V$) and inhibitory ($U$) neurons are shown as the following Equation (\ref{equa_doubleneuorn}):

\begin{equation}
\left\{
\begin{array}{lr}
\vspace{1ex}
\frac{dV_{k,t}}{dt} = g V_{k,t}^2 + m U_{k,t} + I_{k,t} \\
\frac{dU_{k,t}}{dt} = n U_{k,t}
\end{array}
\right.
\text{,}
\label{equa_doubleneuorn}
\end{equation}

where $m$ and $n$ are hyperparameters for describing the inhibition strength and self-dynamics of the inhibitory neuron, respectively. The immediate feedback of spikes from another inhibitory neuron will temporarily block the excitatory neuron's activity for a short time, increasing the scale of dynamics. A schematic diagram depicting the membrane potential under double-neuron dynamics is also shown in Figure \ref{fig_overall}.

% \begin{figure}[htbp]
% \centering
% \includegraphics[width=7cm]{Figure/double_neuron.png}
% \caption{Double-neuron dynamics between excitatory and inhibitory neurons.}
% \label{fig_double}
% \end{figure}

\subsection{Decoder layer}

We use the auto-regressive (AR) decoder with a transformer to post processing the spike trains generated from different types of dynamic neurons. The probability distribution is finally calculated as the whole network output.

\subsection{The global and local learning of DyTr-SNN}

The main training algorithm of DyTr-TNN includes two main parts, i.e., global learning based on the loss function and local learning based on neuronal coding.

We select 3 loss functions for their fitness of the proposed DyTr-SNN. The cross-entropy loss ($L_{CE}$) is used to calculate the probability distribution difference between the model output and the target sequence. A CTC loss $L_{CTC}$ is selected to promote acoustic boundaries' identification accuracy. A quantity loss $L_{Qua}$ is used to limit the differences between the number of boundaries (e.g., the quantity of the target tokens) and the number of spikes (e.g., the number of predicted tokens). Then these three loss functions are integrated, shown as the following Equation (\ref{equa_total_loss}):

\begin{equation}
Loss=\lambda_1 L_{CE} + \lambda_2 L_{CTC} + \lambda_3 L_{Qua}
\text{,}
\label{equa_total_loss}
\end{equation}

where $\lambda_1$, $\lambda_2$, and $\lambda_3$ are hyper-parameters for adjusting the different importance of these three loss functions. After getting $Loss$, we calculate the gradient of all parameters and update our model. For spiking neurons, the gradient approximation trick is used to get around the non-differential feature of the membrane potential containing spikes \cite{RN765}. 

While neuron coding uses the learning ability of a single neuron, which has the complex dynamics mentioned above, the neuron is placed to position token boundaries from the continuous speech in the way of membrane potential accumulation and firing under the encoder-decoder framework. It forwardly integrates the information in the encoded acoustic representations and fires the integrated information to the decoder upon locating a boundary.

\section{Experiments}

The proposed DyTr-SNN was evaluated on a non-toy automatic speech recognition task. The introduction of this dataset, the related experimental configurations, and the experimental results are shown in the following sections.

\subsection{LJSpeech dataset}

The LJSpeech \cite{ljspeech17} was used in this study as a non-toy ASR dataset. It is a 24-hour single-speaker speech dataset comprised of 13,100 short English audio clips with a sample rate of 22,050 Hz. The reference transcriptions are provided for all audio clips, ranging from 1 to 10 seconds.

We split the original LJSpeech dataset into 12,229 samples for training, 348 samples for validation, and 523 samples for testing \cite{ren2020fastspeech}. We first extract the 80-dimension log-Mel spectrogram of these audio clips with 1,024 FFT size, 1,024 window size, and 256 hop length. Then, we apply global cepstral mean and variance normalization (CMVN) to the extracted acoustic features. Finally, we use the processed features as the network input. The corresponding phoneme sequences of LJSpeech audio clips are used as the training target. To generate the phoneme sequences of audio clips, we use the g2pE tool \cite{g2pE2019} to convert transcriptions. The final token set contains 69 phonemes and 4 special tokens: ``sp'', ``spn'', ``eos'', and ``bos''.

% We choose LJSpeech for two reasons: 1) Different from other image datasets, LJSpeech, as a speech-text dataset, contains complex sequence mapping patterns across different temporal scales, which can better highlight the intrinsic ability to process spatio-temporal information formed by our SD-SNN model comprised of transformer modules and spiking neurons with complex dynamics. 2) Compared with other speech datasets such as TIDigits \cite{leonard1984database}, TIMIT \cite{fisher1986darpa}, and LibriSpeech \cite{panayotov2015librispeech}, the audios of LJSpeech are of moderate duration so that we can discriminate enough gap of performances between different neuron dynamics with proper training cost.

\subsection{Experimental setting}

The spiking transformer encoder consists of the convolutional module and the transformer module. The convolutional module has 2 layers followed by a GLU activation function \cite{dauphin2017language}, with each layer using 1-D convolution kernels along the time axis with kernel size 5 and stride 2, and the output channels for them are 512 and 1,024, respectively. The transformer module is comprised of $N=12$ self-attention blocks, sharing the same input dimension ($d_{ff}=2048$), output dimension ($d_{model}=512$), and head numbers ($h=8$). The hidden layer of neuronal dynamics is a 1-D convolution layer with kernel size 3 and 256 output channels. A decoder layer is a stack of 6 self-attention blocks, sharing the same parameters of $d_{model}$, $d_{ff}$, and $h$ as in the encoder layer.

For model training, we apply inverse square root learning rate scheduler and use the Adam optimizer \cite{kingma2014adam} with peak learning rate $5e$-$4$, $\beta_1$=$0.9$, $\beta_2$=$0.98$. $\lambda_1$, $\lambda_2$ and $\lambda_3$ are set to $1.0$, $0.25$, and $1.0$, respectively. We apply dropout $0.2$ to all self-attention blocks and convolution layers for model regularisation. We also apply Specaugment \cite{park2019specaugment} with LD policy to augment acoustic features. For model inference, we use beam search with beam size 5. To measure the performance, we use phoneme error rate (PER) as the metric. We implement our models and methods on Fairseq with PyTorch. The basic implementation of vanilla 1st-order dynamics is available online\footnote{\url{ https://github.com/MingLunHan/CIF-PyTorch}}. 

\begin{figure}[htbp]
\centering
\includegraphics[width=8cm]{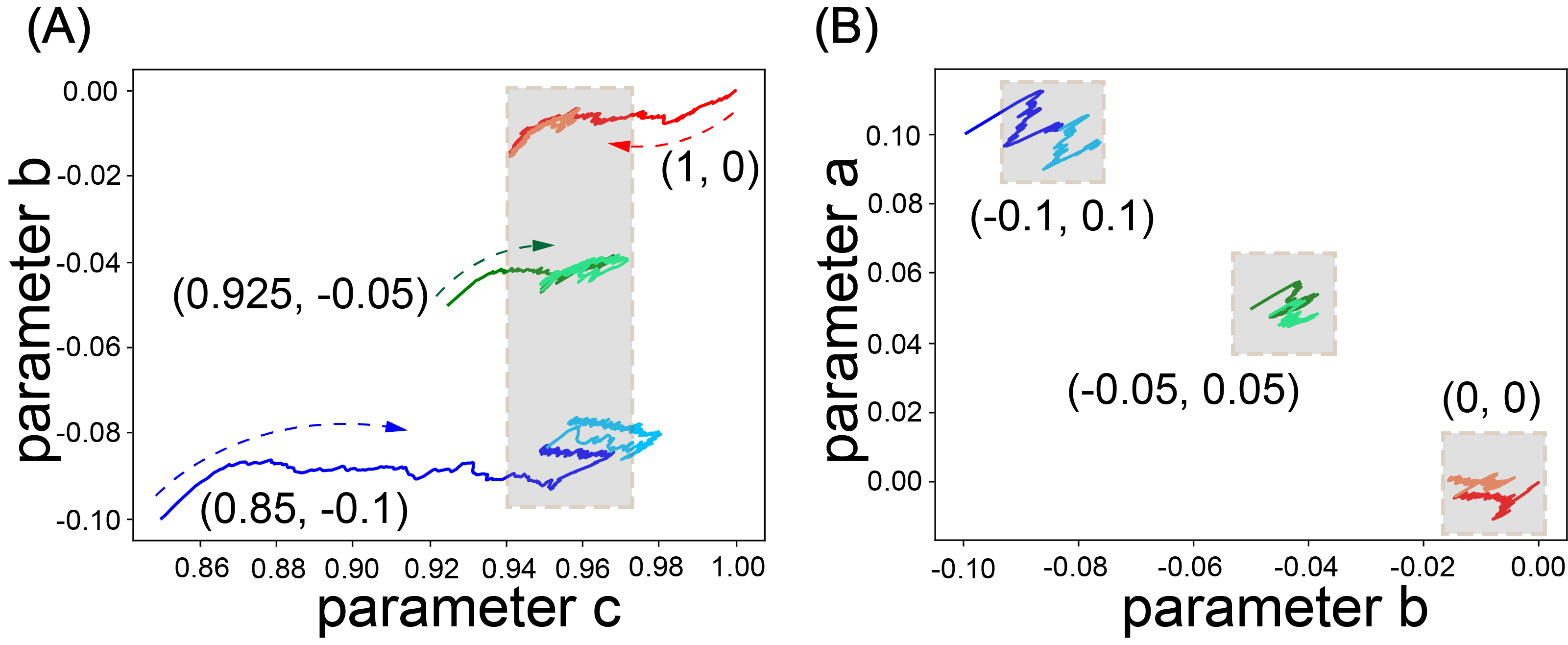}
\caption{Gradually stable trajectories of three parameters. Gray regions are stable areas.}
\label{fig_tra}
\end{figure}

\subsection{Pre-learning of dynamic neurons}

In order to find the proper parameters of DyTr-SNN, we used a mini dataset (1/8 of the original training set) in a smaller network (with a reduced model of 2 transformer encoder layers and 2 decoder layers) first. Then we applied the parameters to the whole dataset for further adjustment.

The parameters for all dynamics followed the same pipeline to get the best configurations. For the 2nd-order dynamics, we first enable the three parameters (i.e., $a$, $b$, and $c$) to be updated along with the synaptic weights learning based on the loss function. During the evolutionary process, these three parameters converged into (0.1014, -0.0832, 0.9506), as shown in the recorded trajectory in Figure \ref{fig_tra}. The $\alpha$ is 0.95, $\Delta h$ is 0.1, $g$ is 0.001, $m$ and $n$ are -0.05, for dynamics in Equations (6) and (8).

% The three initial points (0.85, -0.1, 0.1), (0.925, -0.05, 0.05), (1, 0, 0) are set, and it can be observed from the three trajectories (marked blue, green, and red, respectively) 

\begin{figure}[htbp]
\centering
\includegraphics[width=8cm]{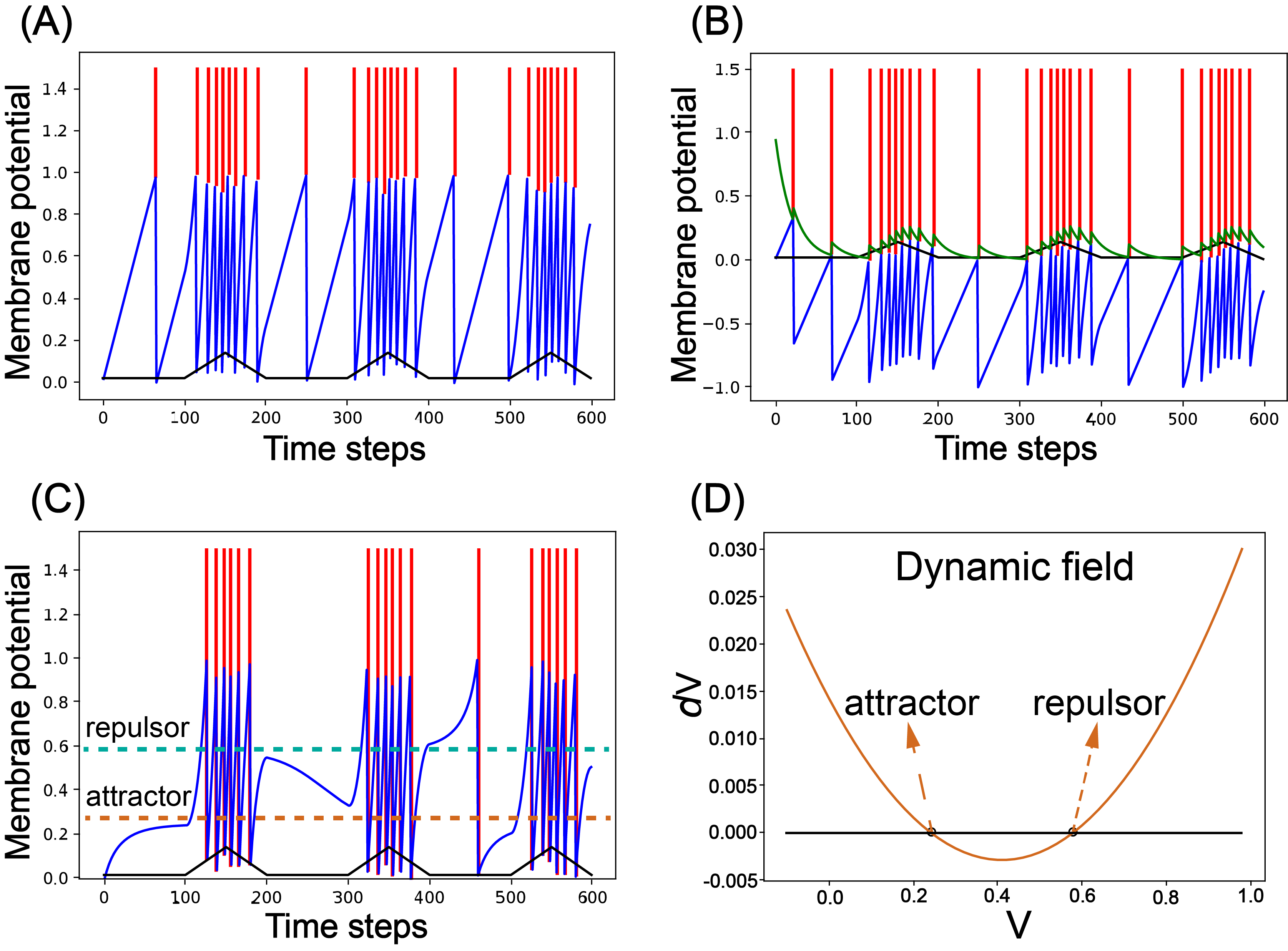}
\caption{Firing patterns of neurons with 1st and 2nd-dynamics. (A) The vanilla 1st-order dynamics. (B) The 1st-order dynamics using dynamic threshold, (C) The 2nd-order dynamics. (D) The diagram depicting the attractor and repulsor is calculated from the 2nd-order dynamics.}
\label{fig_potential}
\end{figure}

After pre-learning, we visualized the firing patterns of these dynamic neurons by giving a simulated current input (e.g., a triangle-linear wave with a minimum value of 0.015 and a maximum value of 0.15). For the vanilla 1st-order dynamics, we found that the number of spikes was linearly updated with the input current (Figure \ref{fig_potential}A). For the 1st-order dynamics using adaptive threshold, we found that the threshold was dynamic and positively correlated with the firing rate, which caused earlier firings (Figure \ref{fig_potential}B). 

\begin{figure}[htbp]
\centering
\includegraphics[width=7cm]{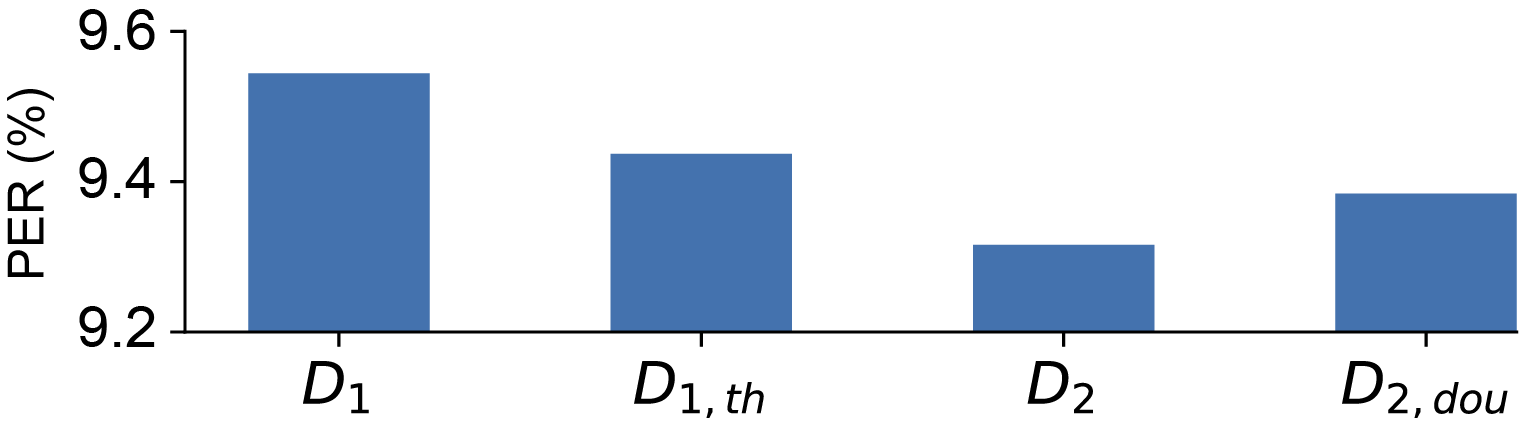}
\caption{Histogram depicting the PER comparison of the proposed DyTr-SNN using different neuronal dynamics and other typical algorithms on the LJSpeech dataset. The PERs for DyTr-SNN using $D_{1}$, $D_{1,th}$, $D_{2}$, and $D_{2,dou}$ were $9.544 \pm0.047$, $9.437 \pm 0.115$, $9.316 \pm 0.029$, $9.384 \pm 0.030$, respectively.
}
\label{fig_PER}
\end{figure}

For the 2nd-order dynamics, we found that the accumulation process of membrane potential was non-linear. The growth rate was not stable even for that given a constant input, caused by the more complex dynamics related to its historical experience (Figure \ref{fig_potential}C). It could be explained by the one attractor and one repulsor in the dynamic process of membrane potential (Figure \ref{fig_potential}D). The attractor means the membrane potential will continuously increase or be leaky, depending on whether it is greater than the attractor. While on the contrary, it will be called a repulsor.

\subsection{Lower phoneme error rate (PER)}

The phoneme error rate (PER) was selected to judge the performance of the proposed DyTr-SNN and other typical algorithms on LJSpeech. Each experiment was repeated three times using initial synaptic weights with different random seeds.

As shown in Figure \ref{fig_PER}, the four dynamics are called $D_1$, $D_{1,th}$, $D_{2}$, $D_{2,dou}$ respectively for convenience. All our DyTr-SNNs using any neuronal dynamics achieved markedly lower PER than the typical algorithm. This result showed the efficiency of the hybrid integration of the spiking transformer and spiking neuronal dynamics. Furthermore, the DyTr-SNN using the 2nd-order dynamics achieved the highest performance than SNNs using the other three types of neuronal dynamics. A possible explanation is that a neuron with more complex dynamics can better catch the complex encoding features from the spiking transformer encoder. The complex encoding represents the transformer-like self-attention encoding, where each frame in the transformer contains the acoustic mutual information of the other frames and its self-information. 

The multi-head attention network can map the same signal vector to multiple spaces, representing complex spatial-temporal information. For example, one phoneme will be embedded in the entire sentence in many ways, depending on how it relates to other phonemes. As a result, the output integrates a variety of possible relationships between the associated acoustic information. Hence, a neuron with more spatio-temporal dynamics can better deal with this situation.

\begin{figure}[htbp]
\centering
\includegraphics[width=7cm]{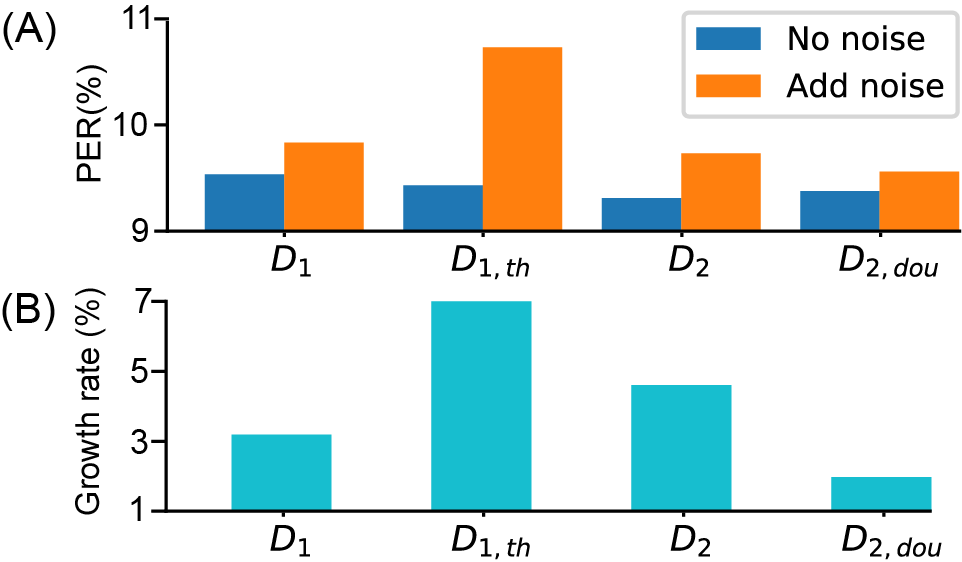}
\caption{Histogram depicting the robustness of DyTr-SNN using different neuronal dynamics. A, The PER before and after adding noises into network input. B, The relative changes of PER before and after giving noises. The PERs for DyTr-SNN using 4 dynamics were $9.748 \pm0.060$, $10.763 \pm 0.043$, $9.745 \pm 0.038$, $9.569 \pm 0.048$, respectively. The relative changes were 3.2\%, 14.1\%,  4.6\%, 2.0\%.}
\label{fig_robust}
\end{figure}

\subsection{Stronger robustness}

The robust computation in SNNs has been reported many times in previous studies. Here we test the robustness of the proposed DyTr-SNN using different neuronal dynamics. We added a uniform noise ranging from 0 to 0.01 to the network input, and the experimental results are shown in Figure \ref{fig_robust}. The performance of networks before and after adding noises showed that different dynamics exhibited a different proportion of PER changes (Figure \ref{fig_robust}A). Furthermore, the relative PER change of DyTr-SNN using $D_{2,dou}$ was the lowest compared to other neuronal dynamics, representing that the $D_{2,dou}$ were more robust than other dynamics.

\begin{figure}[htbp]
\centering
\includegraphics[width=8cm]{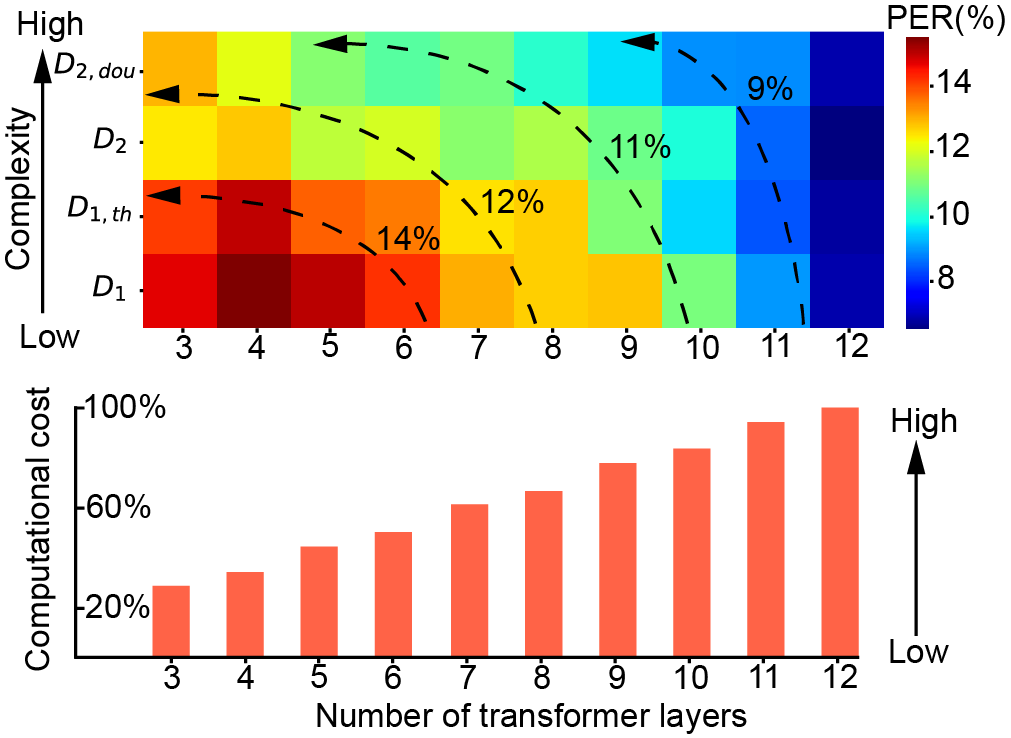}
\caption{PER topography contour measured by different neuronal dynamics (vertical) and the number of layers (horizontal), and computational cost with different model sizes.}
\label{fig_cost}
\end{figure}

\subsection{Lower computational cost}

It is a dilemma that networks with more layers will reach a higher accuracy but, at the same time, take more computational costs. In this study, the proposed DyTr-SNN is energy efficient in the following two aspects: the spiking signals are more efficient than artificial ones caused of their 0/1 feature for efficient computation; the more complex neuronal dynamics will acquire less precoding of information. 

The former has been commonly verified in many related types of research, so we will not further discuss it. Here we focus on the second one and compare a network using the different number of hidden layers and dynamic neurons with different neuronal dynamics. As shown in Figure \ref{fig_cost}, the PER topography contour map showed that the PER is gradually reduced with increased neuronal complexity. A similar conclusion is given using an increasing number of transformer encoder layers. The contour line (marked black) means the models achieved similar performance (e.g., the model with $D_{2,dou}$ and 5 transformer layers is similar to that with $D_1$ and 10 transformer layers in PER). The result shows that the increasing complexity is similar to or equivalent to the increasing encoder layers.

\subsection{Analysis of spikes for acoustic boundary location}

\begin{figure}[htbp]
\centering
\includegraphics[width=7.5cm]{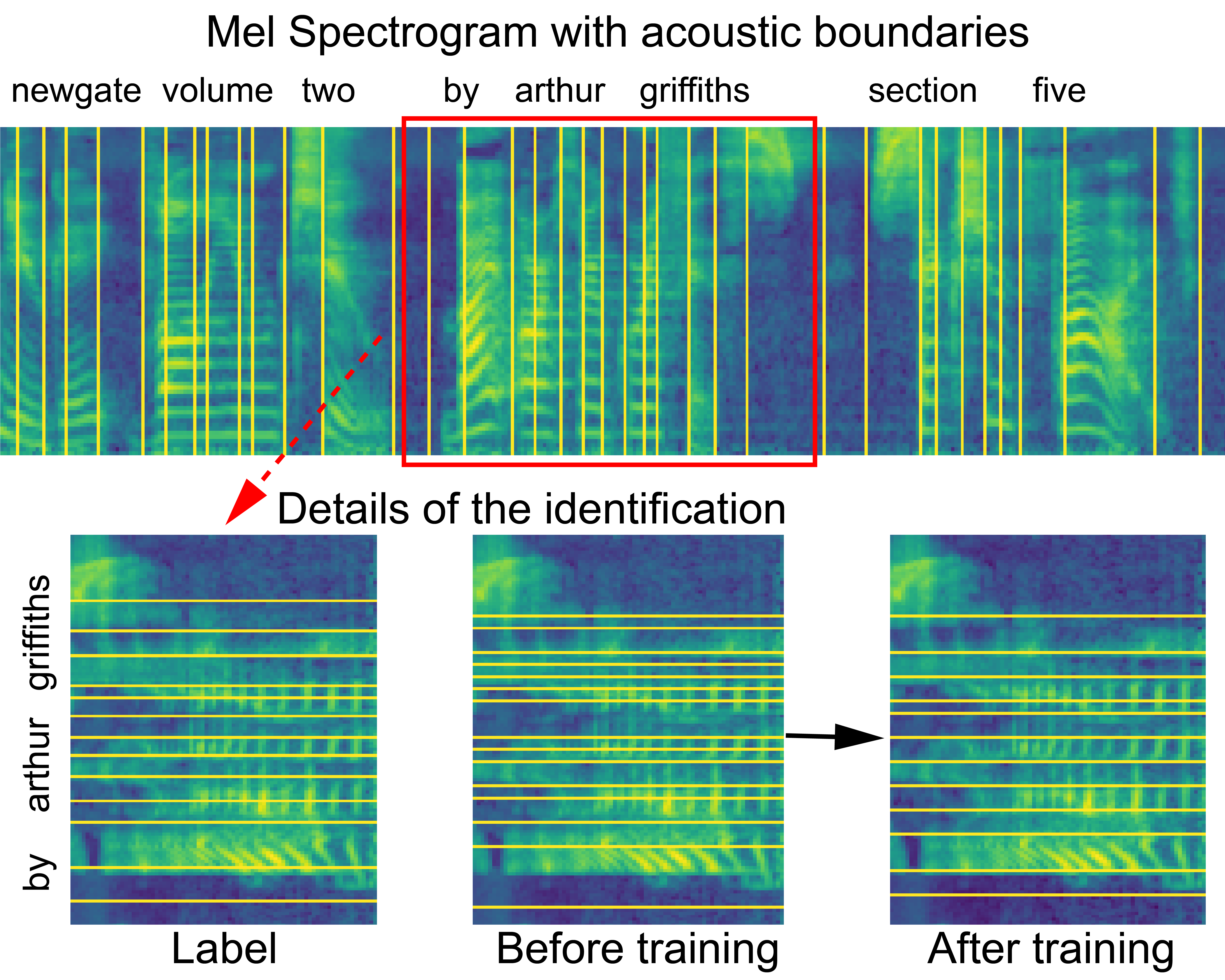}
\caption{The visualization of acoustic boundaries on the Mel-spectrogram at the phoneme level. After training, the boundary locations (marked yellow lines) are closer to the label boundaries than the ones before training.}
\label{fig_analysis}
\end{figure}

The important function of the spiking neuron encoder is to precisely identify the acoustic boundaries at the phoneme level, using spikes of dynamic neurons as the boundary flags. An example of a short speech is shown in Figure \ref{fig_analysis}, which contains a sentence with nine words consisting of 37 phonemes. We select three words to visualize and find that, compared with the ones before learning, the boundary locations after training are closer to the label of boundaries. Here, a more precise positioning means a lower PER and higher performance.

\subsection{Further discussion on hybrid architectures}

Besides combining spiking transformers and dynamic neurons, we run further by integrating artificial transformers and spiking neurons with neuronal dynamics to achieve much higher performance. For spiking transformer, algorithm using $D_1$, $D_{1,th}$, $D_{2}$, and $D_{2,dou}$ achieved PER of $9.544 \pm 0.047$, $9.437 \pm 0.073$, $9.316 \pm 0.029$, and $9.384 \pm 0.030$, respectively. For artificial transformer, algorithm using $D_1$, $D_{1,th}$, $D_{2}$, and $D_{2,dou}$ achieved PER of $6.805 \pm 0.092$, $6.692 \pm 0.115$, $6.455 \pm 0.062$, and $6.786 \pm 0.038$, respectively.

\section{Conclusion}
The spiking neural network (SNN) contains both neuronal dynamics and spiking signals for efficient temporal information processing. In this study, we extend SNN further by using complex neurons inspired by the biological brain and a spiking transformer borrowed from the conventional artificial transformer neural network. Four types of neuronal dynamics are proposed and improved SNNs on the heterogeneity of dynamics, exhibiting the different performance of phoneme error rate, robustness, and computational cost. Further analysis of the linear relationship between the scale of neuronal complexity and the number of transformer layers indicates that the brain uses a relatively shallow neural network with very complex neuronal dynamics for handling different challenging tasks without losing accuracy, efficiency, and robustness. 

\section{Acknowledgments}
This work was supported by the National Key R\&D Program of China (Grant No. 2020AAA0104305), the Strategic Priority Research Program of Chinese Academy of Sciences (Grant No. XDB32070100 and XDA27010404), the Shanghai Municipal Science and Technology Major Project (Grant No. 2021SHZDZX), and the Youth Innovation Promotion Association of CAS.

\bibliography{aaai23}

\begin{thebibliography}{31}
\providecommand{\natexlab}[1]{#1}

\bibitem[{Bellec et~al.(2020)Bellec, Scherr, Subramoney, and et~al.}]{RN557}
Bellec, G.; Scherr, F.; Subramoney, A.; and et~al. 2020.
\newblock A solution to the learning dilemma for recurrent networks of spiking
  neurons.
\newblock \emph{Nat Commun}, 11(1): 3625.

\bibitem[{Cheng et~al.(2020)Cheng, Zhang, Jia, and Xu}]{RN478}
Cheng, X.; Zhang, T.; Jia, S.; and Xu, B. 2020.
\newblock Finite Meta-Dynamic Neurons in Spiking Neural Networks for
  Spatio-temporal Learning.
\newblock \emph{ArXiv}, cs.NE/2010.03140.

\bibitem[{Dauphin et~al.(2017)Dauphin, Fan, Auli, and
  Grangier}]{dauphin2017language}
Dauphin, Y.~N.; Fan, A.; Auli, M.; and Grangier, D. 2017.
\newblock Language modeling with gated convolutional networks.
\newblock In \emph{ICML}, 933--941. PMLR.

\bibitem[{Dong and Xu(2020)}]{linhao2018cif}
Dong, L.; and Xu, B. 2020.
\newblock {CIF:} Continuous Integrate-And-Fire for End-To-End Speech
  Recognition.
\newblock In \emph{{ICASSP}}, 6079--6083. {IEEE}.

\bibitem[{Eliasmith et~al.(2012)Eliasmith, Stewart, Choo, Bekolay, DeWolf,
  Tang, and Rasmussen}]{RN548}
Eliasmith, C.; Stewart, T.~C.; Choo, X.; Bekolay, T.; DeWolf, T.; Tang, Y.; and
  Rasmussen, D. 2012.
\newblock A large-scale model of the functioning brain.
\newblock \emph{Science}, 338(6111): 1202--5.

\bibitem[{Han et~al.(2022)Han, Dong, Liang, Cai, Zhou, Ma, and
  Xu}]{HanDLCZMX22}
Han, M.; Dong, L.; Liang, Z.; Cai, M.; Zhou, S.; Ma, Z.; and Xu, B. 2022.
\newblock Improving End-to-End Contextual Speech Recognition with Fine-Grained
  Contextual Knowledge Selection.
\newblock In \emph{{ICASSP}}, 8532--8536. {IEEE}.

\bibitem[{Han et~al.(2021)Han, Dong, Zhou, and Xu}]{HanDZX21}
Han, M.; Dong, L.; Zhou, S.; and Xu, B. 2021.
\newblock Cif-Based Collaborative Decoding for End-to-End Contextual Speech
  Recognition.
\newblock In \emph{{ICASSP}}, 6528--6532. {IEEE}.

\bibitem[{Ito and Johnson(2017)}]{ljspeech17}
Ito, K.; and Johnson, L. 2017.
\newblock The LJ Speech Dataset.
\newblock \url{https://keithito.com/LJ-Speech-Dataset/}.

\bibitem[{Izhikevich(2003)}]{RN448}
Izhikevich, E.~M. 2003.
\newblock Simple model of spiking neurons.
\newblock \emph{IEEE Transactions on Neural Networks}, 14(6): 1569--72.

\bibitem[{Jia et~al.(2021)Jia, Zhang, Cheng, and et~al.}]{RN766}
Jia, S.; Zhang, T.; Cheng, X.; and et~al. 2021.
\newblock Neuronal-Plasticity and Reward-Propagation Improved Recurrent Spiking
  Neural Networks.
\newblock \emph{Front Neurosci}, 15: 654786.

\bibitem[{Jia et~al.(2022)Jia, Zuo, Zhang, Liu, and Xu}]{RN801}
Jia, S.; Zuo, R.; Zhang, T.; Liu, H.; and Xu, B. 2022.
\newblock Motif-Topology and Reward-Learning Improved Spiking Neural Network
  for Efficient Multi-Sensory Integration.
\newblock In \emph{ICASSP 2022-2022 IEEE International Conference on Acoustics,
  Speech and Signal Processing (ICASSP)}, 8917--8921. IEEE.

\bibitem[{Kingma and Ba(2014)}]{kingma2014adam}
Kingma, D.~P.; and Ba, J. 2014.
\newblock Adam: A method for stochastic optimization.
\newblock \emph{arXiv preprint arXiv:1412.6980}.

\bibitem[{Kugele et~al.(2020)Kugele, Pfeil, Pfeiffer, and Chicca}]{RN307}
Kugele, A.; Pfeil, T.; Pfeiffer, M.; and Chicca, E. 2020.
\newblock Efficient Processing of Spatio-Temporal Data Streams With Spiking
  Neural Networks.
\newblock \emph{Front Neurosci}, 14: 439.

\bibitem[{Maass(1997)}]{RN670}
Maass, W. 1997.
\newblock Networks of spiking neurons: The third generation of neural network
  models.
\newblock \emph{Neural Networks}, 10(9): 1659--1671.

\bibitem[{Mueller et~al.(2021)Mueller, Studenyak, Auge, and
  Knoll}]{mueller2021spiking}
Mueller, E.; Studenyak, V.; Auge, D.; and Knoll, A. 2021.
\newblock Spiking Transformer Networks: A Rate Coded Approach for Processing
  Sequential Data.
\newblock In \emph{2021 7th International Conference on Systems and Informatics
  (ICSAI)}, 1--5. IEEE.

\bibitem[{Park et~al.(2019)Park, Chan, Zhang, Chiu, Zoph, Cubuk, and
  Le}]{park2019specaugment}
Park, D.~S.; Chan, W.; Zhang, Y.; Chiu, C.-C.; Zoph, B.; Cubuk, E.~D.; and Le,
  Q.~V. 2019.
\newblock Specaugment: A simple data augmentation method for automatic speech
  recognition.
\newblock \emph{arXiv preprint arXiv:1904.08779}.

\bibitem[{Park and Kim(2019)}]{g2pE2019}
Park, K.; and Kim, J. 2019.
\newblock g2pE.
\newblock \url{https://github.com/Kyubyong/g2p}.

\bibitem[{Pasqualotto, Dumitru, and Myachykov(2015)}]{RN304}
Pasqualotto, A.; Dumitru, M.~L.; and Myachykov, A. 2015.
\newblock Editorial: Multisensory Integration: Brain, Body, and World.
\newblock \emph{Front Psychol}, 6: 2046.

\bibitem[{Pei et~al.(2019)Pei, Deng, Song, and et~al.}]{RN436}
Pei, J.; Deng, L.; Song, S.; and et~al. 2019.
\newblock Towards artificial general intelligence with hybrid Tianjic chip
  architecture.
\newblock \emph{Nature}, 572(7767): 106--111.

\bibitem[{Perez-Nieves et~al.(2021)Perez-Nieves, Leung, Dragotti, and
  Goodman}]{RN778}
Perez-Nieves, N.; Leung, V. C.~H.; Dragotti, P.~L.; and Goodman, D. F.~M. 2021.
\newblock Neural heterogeneity promotes robust learning.
\newblock \emph{Nat Commun}, 12(1): 5791.

\bibitem[{Posch et~al.(2014)Posch, Serrano-Gotarredona, Linares-Barranco, and
  Delbruck}]{RN308}
Posch, C.; Serrano-Gotarredona, T.; Linares-Barranco, B.; and Delbruck, T.
  2014.
\newblock Retinomorphic Event-Based Vision Sensors: Bioinspired Cameras With
  Spiking Output.
\newblock \emph{Proceedings of the IEEE}, 102(10): 1470--1484.

\bibitem[{Ren et~al.(2020)Ren, Hu, Tan, Qin, Zhao, Zhao, and
  Liu}]{ren2020fastspeech}
Ren, Y.; Hu, C.; Tan, X.; Qin, T.; Zhao, S.; Zhao, Z.; and Liu, T.-Y. 2020.
\newblock Fastspeech 2: Fast and high-quality end-to-end text to speech.
\newblock \emph{arXiv preprint arXiv:2006.04558}.

\bibitem[{Rueckauer et~al.(2017)Rueckauer, Lungu, Hu, Pfeiffer, and
  Liu}]{RN849}
Rueckauer, B.; Lungu, I.~A.; Hu, Y.; Pfeiffer, M.; and Liu, S.~C. 2017.
\newblock Conversion of Continuous-Valued Deep Networks to Efficient
  Event-Driven Networks for Image Classification.
\newblock \emph{Front Neurosci}, 11: 682.

\bibitem[{Stein and Stanford(2008)}]{RN302}
Stein, B.~E.; and Stanford, T.~R. 2008.
\newblock Multisensory integration: current issues from the perspective of the
  single neuron.
\newblock \emph{Nature Reviews Neuroscience}, 9(4): 255--266.

\bibitem[{Tang et~al.(2021)Tang, Kumar, Yoo, and Michmizos}]{tang2021deep}
Tang, G.; Kumar, N.; Yoo, R.; and Michmizos, K. 2021.
\newblock Deep reinforcement learning with population-coded spiking neural
  network for continuous control.
\newblock In \emph{Conference on Robot Learning}, 2016--2029. PMLR.

\bibitem[{Wang et~al.(2018)Wang, Pedretti, Milo, and et~al.}]{RN309}
Wang, W.; Pedretti, G.; Milo, V.; and et~al. 2018.
\newblock Learning of spatiotemporal patterns in a spiking neural network with
  resistive switching synapses.
\newblock \emph{Sci Adv}, 4(9): eaat4752.

\bibitem[{Wei, Han, and Zhang(2021)}]{RN804}
Wei, Q.; Han, L.; and Zhang, T. 2021.
\newblock Spiking Adaptive Dynamic Programming Based on Poisson Process for
  Discrete-Time Nonlinear Systems.
\newblock \emph{IEEE Trans Neural Netw Learn Syst}, PP.

\bibitem[{Zhang et~al.(2022{\natexlab{a}})Zhang, Zhang, Jia, and Xu}]{RN798}
Zhang, D.; Zhang, T.; Jia, S.; and Xu, B. 2022{\natexlab{a}}.
\newblock Multiscale Dynamic Coding improved Spiking Actor Network for
  Reinforcement Learning.
\newblock In \emph{Thirty-Sixth AAAI Conference on Artificial Intelligence}.

\bibitem[{Zhang et~al.(2022{\natexlab{b}})Zhang, Dong, Zhang, and
  et~al.}]{zhang2022spiking1}
Zhang, J.; Dong, B.; Zhang, H.; and et~al. 2022{\natexlab{b}}.
\newblock Spiking Transformers for Event-Based Single Object Tracking.
\newblock In \emph{Proceedings of the IEEE/CVF Conference on Computer Vision
  and Pattern Recognition}, 8801--8810.

\bibitem[{Zhang et~al.(2021{\natexlab{a}})Zhang, Cheng, Jia, Poo, Zeng, and
  Xu}]{RN767}
Zhang, T.; Cheng, X.; Jia, S.; Poo, M.~M.; Zeng, Y.; and Xu, B.
  2021{\natexlab{a}}.
\newblock Self-backpropagation of synaptic modifications elevates the
  efficiency of spiking and artificial neural networks.
\newblock \emph{Sci Adv}, 7(43): eabh0146.

\bibitem[{Zhang et~al.(2021{\natexlab{b}})Zhang, Jia, Cheng, and Xu}]{RN765}
Zhang, T.; Jia, S.; Cheng, X.; and Xu, B. 2021{\natexlab{b}}.
\newblock Tuning Convolutional Spiking Neural Network With Biologically
  Plausible Reward Propagation.
\newblock \emph{IEEE Trans Neural Netw Learn Syst}, PP.

\end{thebibliography}
\end{document}